\documentclass[letterpaper, 10 pt, conference]{ieeeconf}  % Comment this line out if you need a4paper

\IEEEoverridecommandlockouts                              % This command is only needed if
\usepackage{amsmath} % assumes amsmath package installed
\usepackage{amssymb}  % assumes amsmath package installed
\usepackage[usenames, dvipsnames, table]{xcolor}
\usepackage{cite}
\usepackage{multirow}
\usepackage{bm}

\usepackage{caption}
\usepackage{subcaption}

\usepackage{algorithm}     % creates the float "Algorithm 1: My Algorithm"
\usepackage{algorithmic}   % for the algorithm itself, e.g. "for x = 1 to n..."

\makeatletter
\let\NAT@parse\undefined
\makeatother

\usepackage{hyperref}
\usepackage[capitalize]{cleveref}
\usepackage{overpic}
\usepackage{pict2e}

\colorlet{mylinkcolor}{black}
\colorlet{mycitecolor}{black}
\colorlet{myurlcolor}{black}

\hypersetup{
  linkcolor  = mylinkcolor,
  citecolor  = mycitecolor,
  urlcolor   = myurlcolor,
  colorlinks = true,
}

\usepackage[markup=bfit, deletedmarkup=sout, authormarkup=superscript]{changes}
\definechangesauthor[name={ar}, color=teal]{AR}
\definechangesauthor[name={kw}, color=MidnightBlue]{KW}
\definechangesauthor[name={ce}, color=ForestGreen]{CE}
\definechangesauthor[name={td}, color=red]{TODO}
\definechangesauthor[name={ms}, color=orange]{MS}
\definechangesauthor[name={mw}, color=SeaGreen]{MW}

\makeatletter
\let\NAT@parse\undefined
\makeatother

\hypersetup{
  linkcolor  = BrickRed,
  citecolor  = Green,
  urlcolor   = NavyBlue,
  colorlinks = true,
}

\hypersetup{
  linkcolor  = black,
  citecolor  = black,
  urlcolor   = black,
  colorlinks = true,
}

\title{\LARGE \bf Self-Contained Kinematic Calibration of a Novel \\ Whole-Body Artificial Skin for Human-Robot Collaboration}

\author{Kandai Watanabe$^{1}$, Matthew Strong$^{1}$, Mary West$^{1}$,\\ Caleb Escobedo$^{1}$, Ander Aramburu$^{1}$, Krishna Chaitanya Kodur$^{2}$, Alessandro Roncone$^{1}$% stops a space
\thanks{$^{1}$Human Interaction and RObotics (HIRO) Group, Computer Science Department, University of Colorado, Boulder, CO 80309, USA {\tt\small name.surname@colorado.edu}.}%
\thanks{$^{2}$Heracleia Human-Centered Computing Lab, University of Texas at Arlington, Arlington, TX, 76019 USA.}
}
\begin{document}

\maketitle
\thispagestyle{empty}
\pagestyle{empty}

%%%%%%%%%%%%%%%%%%%%%%%%%%%%%%%%%%%%%%%%%%%%%%%%%%%%%%%%%%%%%%%%%%%%%%%%%%%%%%%%
\begin{abstract}
% Why roboskin,

% A robotic, artificial skin can introduce additional capabilities to a state-of-the-art robotic platforms. It grants a sense of touch and an ability to sense nearby objects, leading to an inherently safe system unlike external device measurements (e.g. depth camera) that suffer from overhead calibration, low-bandwidth (usually 30 Hz) and occlusion. However, existing artificial skins are tailored to a specific robot platform and require considerable amount to setup and deploy; especially its pose is unknown after it is randomly mounted on the robot.
% What's our proposal,
In this paper, we present an accelerometer-based kinematic calibration algorithm to accurately estimate the pose of multiple sensor units distributed along a robot body. Our approach is self-contained, can be used on any robot provided with a Denavit-Hartenberg kinematic model, and on any skin equipped with Inertial Measurement Units (IMUs).
To validate the proposed method, we first conduct extensive experimentation in simulation and demonstrate a sub-cm positional error from ground truth data---an improvement of six times with respect to prior work;
subsequently, we then perform a real-world evaluation on a seven degrees-of-freedom collaborative platform.
For this purpose, we additionally introduce a novel design for a stand-alone artificial skin equipped with an IMU for use with the proposed algorithm and a proximity sensor for sensing distance to nearby objects.
In conclusion, in this work, we demonstrate seamless integration between a novel hardware design, an accurate calibration method, and preliminary work on applications: the high positional accuracy effectively enables to locate distributed proximity data and allows for a distributed avoidance controller to safely avoid obstacles and people without the need of additional sensing.
\end{abstract}

%%%%%%%%%%%%%%%%%%%%%%%%%%%%%%%%%%%%%%%%%%%%%%%%%%%%%%%%%%%%%%%%%%%%%%%%%%%%%%%%
%!TEX root = ../root.tex
\section{Introduction}\label{sec:introduction}

% \ms{Suggested section:
% However, as the demand for collaborative robots in increasingly varied environments grows, robots need to be equipped with a variety of sensing capabilities. Additionally, following this increase in collaborative robots in such environments, such sensing capabilities must be calibrated well and easily integrated onto any robotic system, akin to a plug-and-play system. A system with dense sensing capabilities \textit{and} plug-and-play functionality is crucial in such environments.}

% Robots are interacting with humans at an increasingly
% frequent rate both in the workplace and in home environments \cite{bauer2008human}. Historically, the solution to a lack of inherently
% safe robots has been to operate them in a controlled, structured environment. However, interacting with humans often
% requires robots to dynamically adapt to moving obstacles
% and unpredictable interactions. In such cases, it is vital
% that robots can avoid collisions and guarantee safety at
% every moment.
% However,

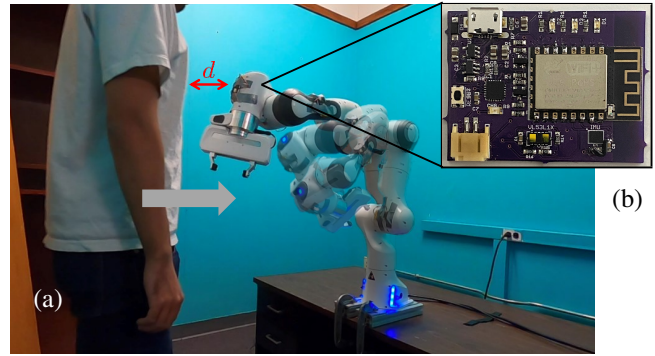
\begin{figure}
    \setlength{\fboxrule}{0.25mm}%
    \makeatletter
    \@wholewidth0.25mm
    \makeatother
    \begin{overpic}[width=0.90\linewidth,percent]{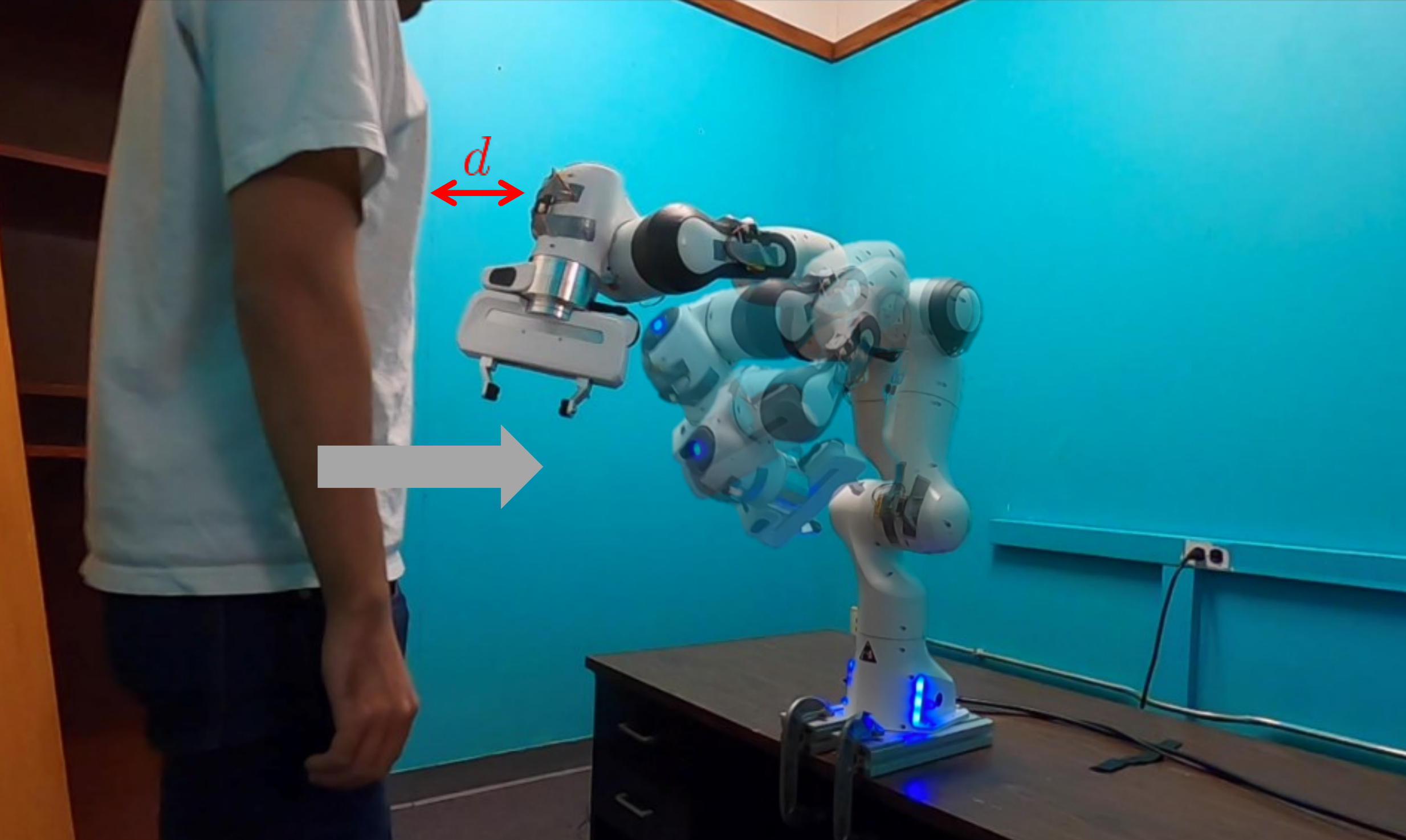}%
        \put(74,32){\frame{\includegraphics[width=0.32\linewidth]{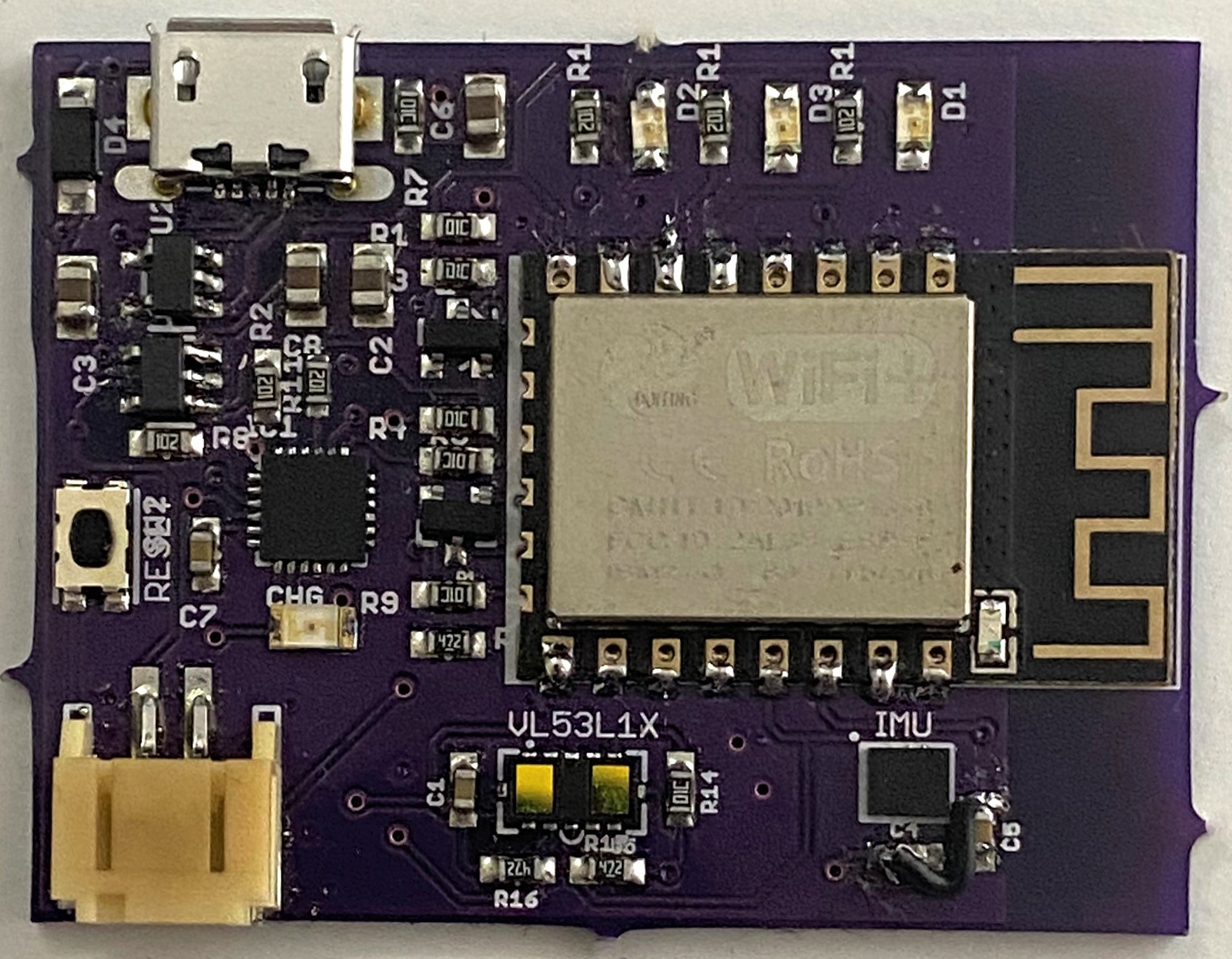}}}%
        \put(4,8){\color{white}(a)}%
        \put(43,46){\linethickness{0.25mm}\color{black}\line(22, 10){31}}
        \put(43,46){\linethickness{0.25mm}\color{black}\line(22,-10){31}}
        \put(103,25){(b)}%
    \end{overpic}
    \caption{We present an algorithm that autonomously calibrates the poses of multiple units of a novel artificial skin prototype mounted on a robot; such poses can then be used to precisely locate sensor data in the robot's frame of reference. a) example of trajectory redirecting for human-aware robot safety: as a person approaches the robot, proximity sensors are activated and the robot successfully avoids the human; b) close-up of the self-contained skin prototype used for validation of the algorithm.}
    \label{fig:panda}
\end{figure}

In recent years, robots have started to leave the structured environments characteristic of factories and laboratories, and they have progressively transitioned to operating with and around people.
However, modern robotic manipulators are limited in their ability to operate in close proximity with humans because they currently lack the necessary sensing capabilities for whole-body perception of humans and obstacles in their immediate surroundings.
Traditionally, collaborative robots designed to work with people (such as the Franka Panda robot in \cref{fig:panda}) resort to either external, sparse, and high-resolution sensing (e.g. RGB-D cameras \cite{flacco2012depth} which have low bandwidth and are prone to occlusions) or on-board, low-resolution, contact-based perception (through e.g. wrist-mounted force/torque sensors, torque sensors on the robot's joints \cite{magrini2015control}, or a combination of the two);
the rest of the body is rarely taken into account.
Under this perspective, distributed \textit{artificial skins} are a promising solution to effectively enable whole-body awareness of contact and information-rich perception of the nearby space, with the goal of robustly perceiving humans and safely operating in close proximity with them \cite{dahiya2012robotic, roncone2015learning, mastrogiovanni2015special, cheng2019comprehensive, metta2010icub, nguyen2018compact,escobedo2021contact}.
However, existing artificial skin technologies are limited by the following:
  i) they often demand a considerable amount of time to design, set up, and deploy the hardware;
 ii) they are usually tailored to a specific platform and cannot easily be ported to different robots;
iii) they generally focus on touch/pressure sensing, and they consequently do not enable perception in the nearby space of a robot;
 iv) they require manual, precise, and time-consuming installation and calibration.
Collectively, these limitations demand a significant overhead, with the net result of limiting research and progress in the field.

In this work, we aim at mitigating these drawbacks by presenting a comprehensive framework for perception, calibration, and robot operation in close proximity with humans.
First, we present an accurate, automated, and self-contained algorithm for \textsl{kinematic calibration} \cite{siciliano2016springer}, which contributes to addressing issue iv).
For the purposes of this paper, we define kinematic calibration as the problem of locating the six-dimensional pose (i.e. position and orientation) of multiple skin sensors distributed along a robot's body.
% A number of previous works have addressed this issue.
% we aim to move away from this paradigm by introducing a robot-independent method that allows for any whole-body skin to be autonomously calibrated and adapted to multiple robot platforms.
%
% We posit that system independent artificial skins have the potential to bring about significant benefits to state-of-the-art robotic platforms.
% In this work, we propose  kinematic calibration algorithm to autonomously localize an arbitrary number of skin units along the kinematic chain of a robot (cf. \cref{sec:calibration}).
Our method does not rely on the use of external metrology systems (such as the laser pointers detailed in \cite{gatla2007automated,hu2012kinematic}), and it exclusively resorts to accelerometer data as read from Inertial Measurement Unit (IMU) sensors conveniently located on the robot skin.
Secondly, we introduce a novel design for a wireless artificial skin that is self-contained, self-powered and capable of proximity and inertial sensing (cf. \cref{fig:panda}).
As detailed in \cref{sec:skin}, our open-source design is robot-independent, it can be readily utilized on a variety of different applications, and it focuses on proximity sensing as the most fundamental feature for a robot to operate around people---thus contributing to issues i), ii), and iii).

The theoretical contribution of the algorithm detailed in \cref{sec:calibration} consists of two main components, which collectively result in improved convergence and greater accuracy if compared to prior work.
First, we introduce a novel formulation of the optimization problem which is now divided in two distinct steps---orientation estimation and subsequent position estimation.
%We divide the original optimization problem into two steps and motivates quicker convergence as detailed in \cref{sec:results}.
Second, acceleration-based calibration of the skin is simplified without loss of accuracy to further accelerate convergence.
%An accurate estimation of SU acceleration is required to compute an accurate error function between the estimated and measured acceleration values. This error function is minimized to optimize the estimated SU pose.
%
The proposed method is initially evaluated in simulation and subsequently validated on a real robot platform (see \cref{sec:experiment}); collectively, our results demonstrate that our contribution significantly outperforms existing state-of-the-art approaches in terms of convergence accuracy, thus bringing us one step closer to a future of plug-and-play and multi-functional sensing.
%Additionally, we demonstrate accuracy and versatility by performing multiple calibrations using different SU poses, with the accuracy demonstrating its versatility.
Finally, the effectiveness of our holistic approach is demonstrated with a minimal obstacle avoidance controller (similarly to \cite{flacco2012depth, flacco2015depth,roncone2016peripersonal,roncone2015learning,nguyen2018compact,jain2013reaching, yuan2017gelsight, yamaguchi2017implementing}) in which calibrated proximity readings are used to safely operate around people (\cref{sec:results}).
Our prototype interaction effectively demonstrates how on-board, distributed, high-bandwidth proximity sensing constitutes a promising direction for future work in close-proximity Human-Robot Collaboration.

\section{Background and related work}\label{sec:background}

% \ms{Suggested section:
% However, as the demand for collaborative robots in increasingly varied environments grows, robots need to be equipped with a variety of sensing capabilities. Additionally, following this increase in collaborative robots in such environments, such sensing capabilities must be calibrated well and easily integrated onto any robotic system, akin to a plug-and-play system. A system with dense sensing capabilities \textit{and} plug-and-play functionality is crucial in such environments.}

% Robots are interacting with humans at an increasingly
% frequent rate both in the workplace and in home environments \cite{bauer2008human}. Historically, the solution to a lack of inherently
% safe robots has been to operate them in a controlled, structured environment. However, interacting with humans often
% requires robots to dynamically adapt to moving obstacles
% and unpredictable interactions. In such cases, it is vital
% that robots can avoid collisions and guarantee safety at
% every moment.
% However,

In the following, we briefly detail prior research relevant to the purposes of this work, and we clarify how our work is positioned with respect to prior efforts.

\paragraph*{A. Kinematic calibration}

Calibration of a robot's kinematic chain is a core problem for industrial and collaborative robotics, as precise modeling of a robot's kinematics and dynamics has direct impact on performance and operation of a robot manipulator.
For this reason, kinematic calibration has been extensively investigated in past decades, and a number of techniques have been devised for both \textsl{open-loop} and \textsl{closed-loop} approaches (please refer to \cite{hollerbach2016model} for an overview).
However, these works are different in nature from the proposed method in that they are improving an already available kinematic model---which is a significant prior that facilitates convergence of the optimization and accuracy of the results.
In our work, we are concerned with kinematically calibrating artificial skin sensors that are randomly placed on a robot's surface, i.e. without any prior information to rely upon.
Of particular relevance to this paper is the work performed in the area of calibration of robot kinematics based on IMU data \cite{du2013imu}, which overlaps with the work in human body estimation for motion capture, e.g. \cite{roetenberg2009xsens,zhu2004real}.
However, in addition to dealing with a different problem (calibration of an existing kinematic model vs calibration of a free-moving skin unit), existing approaches typically rely on sensor fusion techniques that integrate information coming from a variety of different sensors (e.g. magnetometers, GPS, ultra-wideband radio, cameras, and more). In our work, we exclusively employ accelerometer data as we are constrained by the compact hardware prototype we have introduced.

\paragraph*{B. Artificial skins for robotics}

In recent years, there has been steady and significant progress on developing technologies for distributed artificial sensing in robotics. While the majority of prior work focuses on pressure and touch sensing for robotic grippers to enhance manipulation and grasping (e.g. \cite{yamaguchi2017implementing,yuan2017gelsight,patel2016integrated}), a parallel line of research has aimed attention at whole-body, dense coverage of a robot's surface \cite{metta2010icub,hughes2018robotic,cheng2019comprehensive,mastrogiovanni2015special}.
Our paper capitalizes on this body of work, and improves it in two directions:
 i) by focusing on prior-to-contact perception of the nearby space of a robot rather than on-contact, touch-based perception (see also \cite{hughes2018robotic,goger2013tactile}), which is an under-explored area of interest;
ii) by developing a plug-and-play, self-contained hardware element that can be replicated at scale and utilized to cover a variety of different robots with minimal overhead.
Our work is based on the belief that artificial skins should not be limited to few sensing capabilities but rather house a variety of features. To our knowledge, this is the first attempt at providing such degree of modularity and flexibility.

\paragraph*{C. Kinematic calibration of artificial skins}

To realize safe, robust and reliable obstacle detection, it is of paramount importance to precisely locate the relative pose of each sensor with respect to the kinematic chain of the robot. As mentioned in \cref{sec:background}.A, this process is not to be confused with calibration of a robot's kinematic model, as the sensors to be calibrated can be placed anywhere on the robot body.
Traditionally, kinematic calibration of artificial skin is a procedure that is manually performed by the robot operator, and only recent work has started to automate it.
% A number of previous works have addressed this issue.
\cite{roncone2014automatic,stepanova2019robot} presented a method where the iCub humanoid robot \cite{metta2010icub} performed ``self-touch'' actions on its robotic skin, allowing for an observation of the 3D position of the end-effector and computation of its Denavit-Hartenberg [DH] parameters \cite{hartenberg1955kinematic}. While this approach is autonomous and does not rely on external measurements, the iCub’s bi-manual self-touch capabilities would not be possible on commercially-available collaborative manipulators due a lower number of Degrees of Freedom (DoFs) and consequently reduced manipulability.
Of more generalizability is the work detailed in  \cite{mittendorfer2012open}, where tactile sensors and kinematic information of the robot arm were estimated using acceleration data collected from a set of IMUs mounted within the skin.
However, while this work serves as reference for the method detailed in this paper, its average real-world positional error renders it ineffective in practice---as detailed in \cref{sec:results}.

\section{Novel Design of Artificial Skin}\label{sec:skin}

In the following, we introduce a modular design for a whole-body artificial skin with dynamic sensing capabilities.
% AR: we need to say somewhere what we mean with SU, so I put the text back
We define a \textsl{skin unit} (SU) as the most minimal, atomic skin element needed to perform autonomous kinematic calibration and nearby-space perception; for the purposes of the algorithm detailed in \cref{sec:calibration}, the SU shown in \cref{fig:panda}b is composed of an inertial measurement unit (IMU) and a proximity sensor; contact sensing can be achieved through a combination of accelerometer and gyroscope data.
%
% Future iterations of the hardware will include heterogeneous sensing such as pressure, temperature, and human presence sensing.
%
The proposed skin is characterized by off-the-shelf sensors that enable low overall cost (approximately $\$36$) and low power consumption (with an operational range of $110-160$mA), which effectively allows for a self-contained package to ease prototyping and deployment.
The SUs used in this work (see \cref{fig:estimated_imus} for reference) operate with external $100mAh$ batteries and can operate continuously for up to $10$ hours.

% package  and wireless communication for fast prototyping and deployment; it is independently powered by means of a

% It is a wireless, stand-alone design equipped with an FCC certified WiFi enabled ESP8266 microcontroller, LSM6DS3 IMU, VL53L1X time-of-flight distance sensor and integrated LiPo battery charging.

% We have envisioned a low-cost, low-power, skin unit capable of tactile sensing by measuring a robot's linear acceleration and angular velocity as well as physical interaction and the surrounding environment up to four meters from the skin mount location.

\paragraph*{Sensor Selection and Electronics Considerations}

The $33 \times 36$mm SU prototype detailed in \cref{fig:panda}b is a stand-alone design equipped with a WiFi-enabled ESP8266 microcontroller, a USB-to-UART bridge for programming, visual-based hardware debugging, and several system status LEDs.
The SU is a $2$-layer stack-up rigid PCB fabricated on $0.15$mm Kingboard copper clad laminate $175Tg FR4$ substrate, and it is powered by means of an integrated Lithium-Polymer battery (for an operation time of $7$ to $10$ hours on a single charge).
We selected an LSM6DS3 iNEMO inertial measurement unit, which is characterized by a full-scale acceleration range of $\pm 2/\pm 4/\pm 8/\pm 16 g$, an angular rate range of $\pm 125/\pm 245/\pm 500/\pm 1000/\pm 2000$ degrees per second, and robustness against mechanical shock.
Proximity sensing is provided by a VL53L1X time-of-flight (ToF) sensor, which was chosen for its accuracy, ranging distance (up to $4$m), field of view ($27^\circ$), ranging frequency of $50$ Hz, I2C communication protocol, and small package.
Both sensors were also chosen for their high bandwidth: with six SUs mounted on the robot (\cref{fig:panda}) and sending data wirelessly, we were able to receive data at a rate of $100$Hz from the IMUs and $50$Hz from the distance sensor (although the sensor is capable of a nominal bandwidth of $100$Hz).
All files related to the design can be found at the associated GitHub repository\footnote{\ Finalized circuit schematics can be seen at \url{https://github.com/HIRO-group/RoboSkin_Circuit_Schematics}.}.

\section{Accelerometer-based kinematic calibration}\label{sec:calibration}

\begin{figure}
    \centering
    \includegraphics[width=\linewidth]{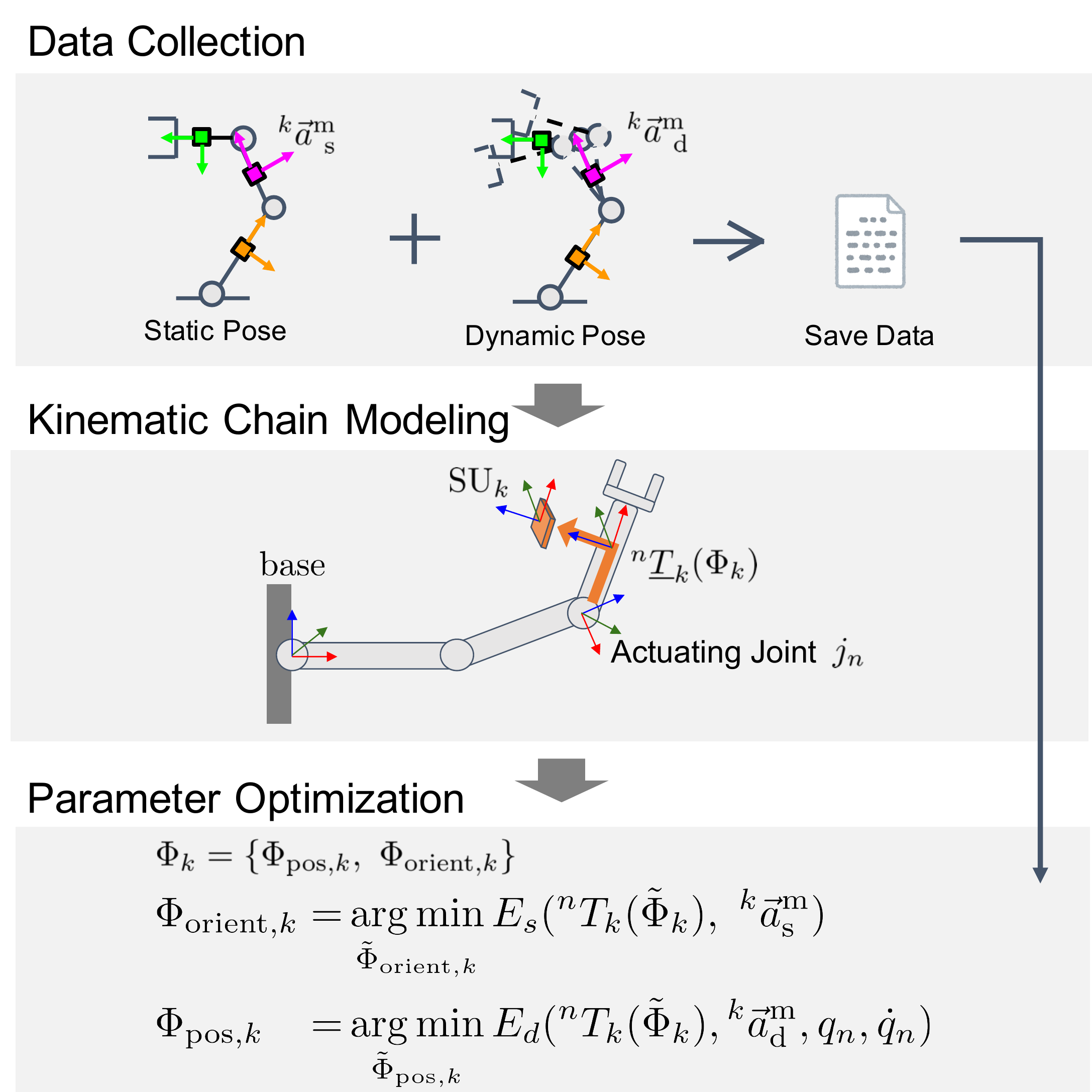}
    \caption{Overview of the kinematic calibration algorithm:
    1) collect IMU data ${}^{k}\vec{a}^{\text{m}}_{\ \text{s}} $ and $ {}^{k}\vec{a}^{\text{m}}_{\ \text{d}} $ at static and dynamic poses for all SU$_k, \ \forall k = \{1, \cdots, K \}$;
    2) define a kinematic model from joint $j_n$ to SU$_k$ expressed as ${}^{n}T_{k}$ that is parameterized by $\Phi_k$;
    3) optimize the parameters $\Phi_k$ by minimizing the static and dynamic error functions $E_s$ and $E_d$ for each SU$_k$.
    $q_n$ and $\dot{q}_n$ represent joint angle and velocity of $j_n$.}
    \label{fig:kinematic_calibration_overview}
\end{figure}

% In the next step, kinematic chain modeling, we define a kinematic chain model of the robot that designates the connectivity between joints and SUs using transformation matrices. For example, in  \cref{fig:kinematic_calibration_overview}, the transformation from a reference segment (RS), the base to the $i$th SU can be expressed as ${}^{RS}\underline{T}_{SU_i}$. In most cases, Denavit-Hartenberg (DH) parameters between joints ${}^{RS}\underline{T}_{DoF_i}$ are provided by the manufacturer. However, the DH parameters that define the connectivity between joints and SUs are unknown (${}^{DoF_i}\underline{T}_{SU_i}$ in \cref{fig:kinematic_calibration_overview}: Modeling). Finally, the parameter optimization step estimates the DH parameters by minimizing the errors between the measured and estimated IMU accelerations (Refer to the equation in \cref{fig:kinematic_calibration_overview}: Parameters Optimization). Acceleration of each IMU ${}^{SU_i} \vec{a}_{p,i,d}^{measure}$ can be predicted by using the kinematic chain model (on the left) with the current estimated DH parameters $\tilde{\theta}_{SU_k}, \tilde{d}_{SU_k}, \tilde{\theta}, \tilde{d}, \tilde{\alpha}, \tilde{a}$. In our work, the DH parameters are sequentially optimized from the base to the end-effector, step by step. In this section, we detail each step, in depth, of the calibration algorithm used to automatically locate and orient an arbitrary number of SUs.

The goal of kinematically calibrating SUs is to identify the $6$D poses (i.e, $3$D positions and $3$D orientations) of each SU mounted on the robot.
% }{Kinematic calibration is the process of estimating the parameters that make up the kinematic structure of a robot.
% The objective of kinematically calibrating SUs is to identify the $6$D poses (i.e, $3$D positions and $3$D orientations) of each SU mounted on the robot so that the sensor measurement data can be precisely located with respect to the robot's kinematic chain and frames of reference.}
With a known pose, the robot can make better sense of tactile information; in \cref{fig:panda,sec:results}, we use this information to demonstrate how an obstacle-avoidance controller can utilize proximity sensor data to detect objects at a distance (up to four meters) and alter the robot's trajectory to prevent collisions.

In this work, we formalize the problem of estimating SU poses along a robot arm with the Denavit-Hartenberg convention detailed in \cite{siciliano2016springer}.
Importantly, our novel kinematic calibration algorithm is exclusively based on accelerometer readings, as our preliminary testing has indicated that angular velocity readings from the gyroscope introduce significant noise and drift to the data and do not improve performance.
We accomplish this by collecting three-axis linear acceleration data $\left(a_x, a_y, a_z\right)$ from each SU in multiple robot configurations.
%
% commented out, was causing errors
% \deleted{Algorithm validation and evaluation (see \cref{results}) are primarily performed in simulation with Gazebo \cite{koenig2004design}, and then ported to a real-world Franka Emika Panda robot. We demonstrate that our algorithm outperforms the state-of-the-art \cite{mittendorfer2012open} and allows for precise autonomous calibration on 7-DoF robotic arms in both simulated and real-world environments.}
%
\cref{fig:kinematic_calibration_overview} shows a high-level overview of the proposed kinematic calibration algorithm; its three core elements are detailed below.

\subsection{Data Collection and Generation}\label{sub:data-collection}

The data collection step is tasked with obtaining sensor readings used in the batch optimization algorithm detailed in \cref{sub:optimization}.
Data collection is split in two parts: static-pose data collection and dynamic-pose data collection. For the purposes of this work, the order in which these steps are performed is not relevant.
The static-pose step collects accelerometer readings in multiple joint configurations with a stationary robot, which serves as a baseline to compensate for static forces acting on the sensor when the robot is motionless---i.e., gravity and baseline measurements.
The dynamic-pose step is tasked with collecting data with each individual joint oscillating in a sinusoidal pattern to span its full operational range; such information is used to perform the optimization detailed in \cref{sub:optimization} after the baseline measurements are compensated for.
More specifically, we independently actuate each joint $p$ with a low-level velocity control defined by $\dot{q}_{p} = A\sin(2\pi ft)$, where $f$ is the frequency, $t$ is the time, $\dot{q}_{p}$ is the joint velocity, and $A$ is the amplitude of the pattern.
This formulation allows to flexibly and conveniently reach either of joint angle, velocity or acceleration limit; %advantages of employing sinusoidal pattern are that it is easy to compute angular position, velocity and acceleration and the function is smooth so that it translates well onto the real robot.
% More specifically, we actuate each and every joint with a sinusoidal velocity $\dot{q}_{p}$ in order for the joint to span its full operational range.
%
% Regarding static data collection, we take a static measurement in order to remove those forces when calculating the amount of acceleration the sensor feels. Then, we move the reference joint in a sinusoidal pattern at each position and save the acceleration throughout this process.
%
%During this joint actuation, we employ a sinusoidal pattern in order to exert a large acceleration at each joint angle.
in particular, exerting large acceleration allows for a larger signal-to-noise ratio, thus helping the optimizer identify acceleration values in the presence of noise.
The data collection process is performed once and saved for later use.
%
% \begin{align}
%     \label{eq:sinusoidal_motion}
%          {q}_{p} &= \frac{A}{2\pi f} (1 - \cos(2\pi ft) \nonumber \\
%      \dot{q}_{p} &= A\sin(2\pi ft)                                \\
%     \ddot{q}_{p} &= 2\pi fA\sin(2\pi ft)                \nonumber
% \end{align}
% \begin{equation}
%     \label{eq:sinusoidal_motion}
%     \dot{q}_{p} &= A\sin(2\pi ft)
% \end{equation}
% %

\subsection{Kinematic Model Considerations}\label{sub:kinematic-model}

\begin{figure}
    \centering
    \includegraphics[width=8cm]{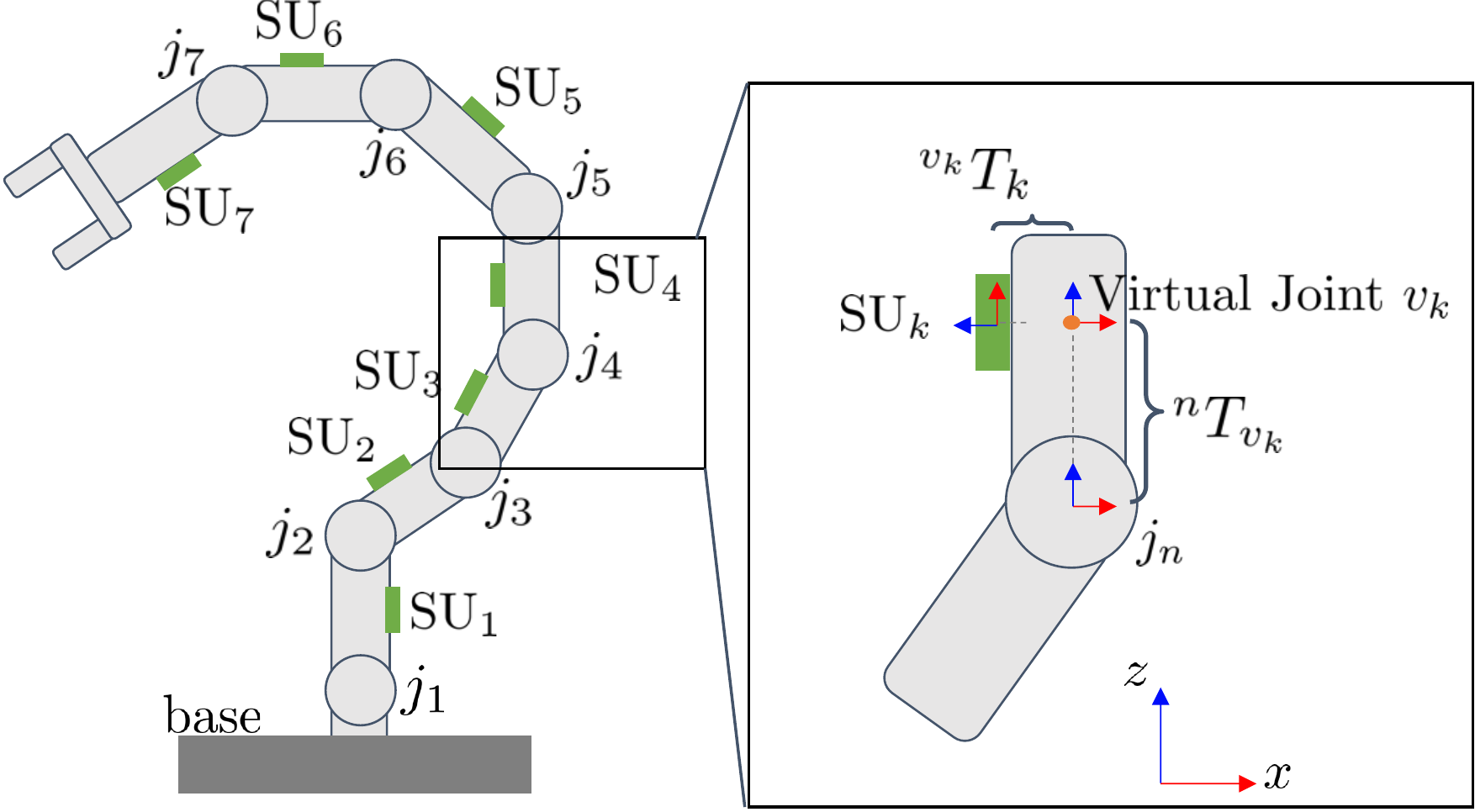}
    \caption{Schematic depiction of SUs mounted on a robotic arm (left) with each kinematic element of the robot labeled. The zoomed-in image on the right illustrates how the SU poses are estimated through DH parameters of the joint connecting to the previous link: the transformation from the $n$-th joint to the $k$-th SU can be computed via an intermediate ``virtual joint'' $v_{k}$ located perpendicular to both the joint's and the SU's reference frames. The axes depict the three coordinate systems: joint $n$, SU $k$, and corresponding virtual joint $v_{k}$.}
    \label{fig:mounted_imu}
\end{figure}

Before running the optimization step detailed in \cref{sub:optimization}, it is imperative to define a suitable kinematic model of the robot.
In this paper, we employ the modified Denavit-Hartenberg representation detailed in \cite{hollerbach2016model}, which is based on the original work in \cite{hartenberg1955kinematic}.
% to construct a kinematic model that relates the pose of each SU with respect to the reference segment (RS) of the link it is mounted on---the gray rectangle at the bottom left in \cref{fig:mounted_imu}.
The DH notation is particularly convenient because it allows for expression of relative poses of kinematic elements along a robot's chain with only four parameters rather than six, thus reducing the dimensionality of the optimization problem.
This dimensionality reduction is achieved by leveraging the standard mechanics of prismatic and revolute joints used in the vast majority of robot arms; however, the same cannot be said of skin units, which can be placed arbitrarily along a robot arm and cannot take advantage of this simplification.
To guarantee compatibility with the DH convention, a ``virtual joint'' is located between joint $n$ and SU $k$, as depicted in \cref{fig:mounted_imu}. This conveniently allows for the problem of estimating the $6$ degree-of-freedom (DoF) transformation matrix ${}^{n}{T}_{k}$ between the $n$th joint's reference frame and the $k$th SU's frame to be decomposed into the estimation of two, DH-compatible, $4$-DoF matrices:
\begin{equation}
    \small
    {}^{n}{T}_{k} = \Bigl[
    \begin{array}{cc}
        {}^{n}R_{k} & {}^{n}\vec{r}_{k} \\
        0 & 1
    \end{array} \Bigr] = {}^{n}{T}_{v_{k}}\left[ \Phi \left( v_k \right) \right] \cdot {}^{v_{k}}{T}_{k} \left[ \Phi \left( SU_k \right) \right] ,
    % \label{eq:Tnk}
    \label{eq:transformation_matrix_decomposition}
\end{equation}
where ${}^{n}R_{k}$ and ${}^{n}\vec{r}_{k}$ are the rotational and translational components of ${}^{n}{T}_{k}$, and $\Phi_k = \left[ \Phi \left( v_k \right), \Phi \left( SU_k \right) \right]$ are the DH-compatible parameters to be estimated---composed of $\Phi \left( v_k \right) = \left(d_{v_k}, \theta_{v_k}, 0, 0\right)$ for virtual joint $v_k$, and  $\Phi \left( SU_k \right) = \left(d_{SU_k}, \theta_{SU_k}, a_{SU_k}, \alpha_{SU_k}\right)$ for the corresponding $SU_k$, where $d$, $\theta$, $a$ and $\alpha$ represent displacement along z-axis, rotation around z-axis, displacement along x-axis and rotation along x-axis of the local frame respectively (See \cref{fig:mounted_imu}).
Note that the above equation still represents a $6$-DoF problem, thus flexibly allowing for any SU pose to be represented relative to its corresponding joint $n$.

\subsection{Parameter Optimization}\label{sub:optimization}

% \begin{figure}
%     \centering
%     \includegraphics[width=0.8\linewidth]{imgs/OptimizationSteps.pdf}
%     \caption{An overview of the Kinematic Calibration Process. The optimization is done for each SU$_k$ step by step for all $K$ SUs from reference segment (RS) to the end-effector.}
%     \label{fig:optimization_process}
% \end{figure}

The last step of the optimization algorithm pertains with parameter optimization, which is tasked with calibrating the kinematic model detailed in \cref{sub:kinematic-model} is with the data collected in \cref{sub:data-collection}.
Initially, the DH parameters for all SUs are randomly initialized within a given range (cf. \cref{sec:experiment}).
At every iteration step, the ground truth accelerations are estimated from the kinematic model as detailed in \cref{subsub:accel-estimation};
then, a global optimization algorithm is used to estimate the DH parameters by minimizing the error between actual and predicted accelerations.
This sequence iterates until it reaches a stopping condition; that is, when the iterative update in DH parameters estimations becomes sufficiently small. Details are provided below.

\subsubsection{Acceleration Estimation} \label{subsub:accel-estimation}

The acceleration ${}^{k}\vec{a}_k$ exerted on SU$_k$ in its frame of reference (FoR) $k$ can be estimated via the Newtonian Equation of motion for a rotating coordinate system.
More specifically, the total acceleration can be seen as a composition of three components: gravitational acceleration $\vec{g}_k$, centripetal acceleration $\vec{a}_{\text{cen},k}$ and tangential acceleration $\vec{a}_{\text{tan},k}$.
% \ar{fix this sentence}Centripetal acceleration is the acceleration the skin unit as a joint $j_n$ rotates with velocity $\dot{q}_{n}$ and the tangential acceleration is the acceleration the skin unit measures perpendicular to its centripetal acceleration.
These three components can be formalized as follows:
%
% \begin{itemize}
%     \item why the acceleration does not depend on $\ddot{q}$? ... It does depend on $\ddot{q}$, but I forgot to include that in the first equation. Please closely look at the third equation.
%     \item Why does it depend on $q$? ... It is used to compute Rotation Matrix $R$
%     \item Why $\vec{\dot{q}}_{n}$ is a vector? ... Because it is rotated around z axis of the joint coordinate. The vector representation is needed to rotate an vector $r$ in 2nd equation
%     \item Why is there a ${}^{k}R_{n}$ and a ${}^{n}\vec{r}_{k}$ (with n and k inverted)? ... Because ${}^{k}R_{n}$ is used to rotate ${}^{n}\vec{r}_{k}$ from coordinate n to coordinate k. The left super script denotes which coordinate you are / will be in.
%     \item What is ${}^{k}R_{n}$? ... It's a rotation matrix that rotates from coordinate n to k.
%     \item Why ${}^{n}\vec{r}_{k}$ is not called ${}^{n}\vec{p}_{k}$ given that it is a position? ... It can but why can't it be $r$? It's a vector.
% \end{itemize}
%
%
\begin{align}
\small
^{k}\vec{a}_k (\Phi_k, q_n, \dot{q}_n, \ddot{q}_n) &= {}^{k}R_{n} (\Phi_k, q) \cdot\Bigl(\vec{g}_k + \vec{a}_{\text{cen},k} + \vec{a}_{\text{tan},k} \Bigr)
\label{eq:total-acceleration}\\
\vec{a}_{\text{cen},k}(\Phi_k, q_n, \dot{q}_n) &= \vec{\dot{q}}_{n} \times\left(\vec{\dot{q}}_{n} \times {}^{n}\vec{r}_k (\Phi_k, q) \right) \label{eq:centripetal-acceleration}\\
\vec{a}_{\text{tan},k} (\Phi_k, q_n, \ddot{q}_n) &= \vec{\ddot{q}}_{n} \times {}^{n}\vec{r}_k (\Phi_k, q) \label{eq:tangential-acceleration}
\end{align}
With reference to \cref{eq:centripetal-acceleration}, the centripetal acceleration $\vec{a}_{\text{cen},k}$ is composed of the joint angular velocity $\vec{\dot{q}}_{n}=[0, 0, \dot{q}_n]$ (which can be measured during data collection), and the position vector ${}^{n}\vec{r}_{k}$ which represents the translational component of ${}^{n}{T}_{k}$ in \cref{eq:transformation_matrix_decomposition}.
Since they are all defined in the $n$th rotating joint's reference frame, it has to be rotated from reference frame $j_n$ to SU$_k$ through ${}^{k}R_{n}$. Note that the full acceleration vector calculation is parameterized by $\Phi_k$, which are the to-be-optimized DH parameters.

\begin{figure}
    \centering
    \includegraphics[width=\linewidth]{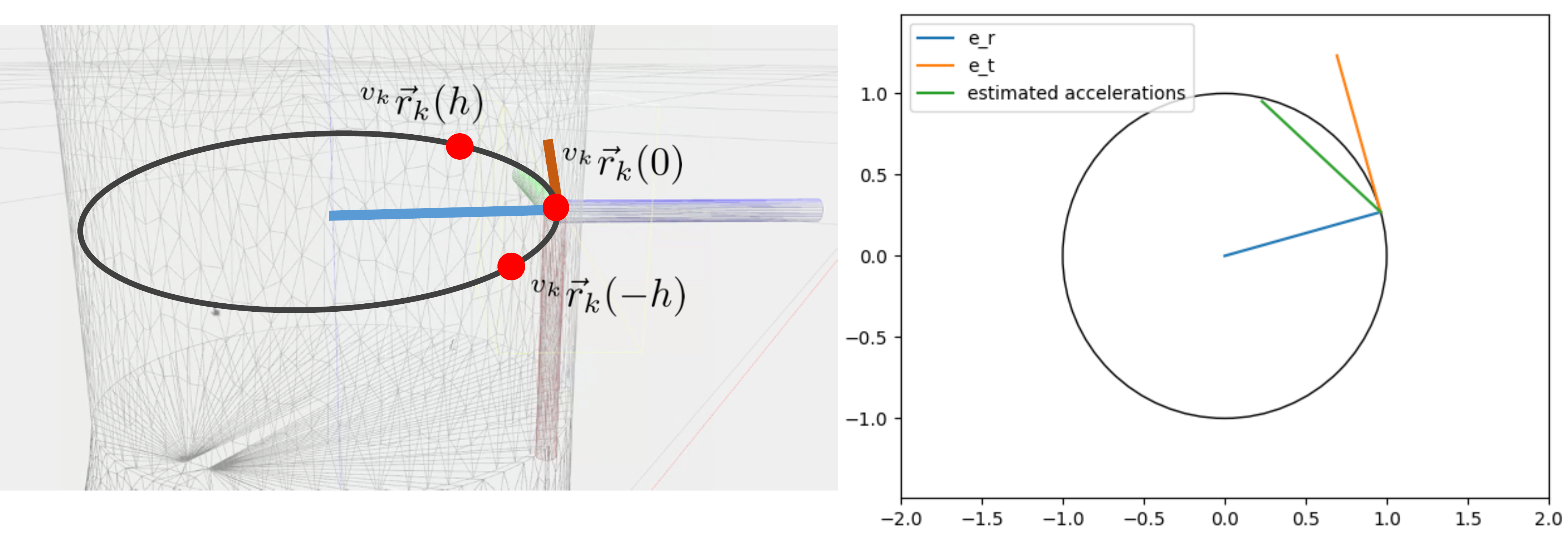}
    \caption{Illustration of the numerically computed tangential accelerations for joint $n$ in its respective FoR. The plot on the right shows the accelerations computed by taking the second derivative of the position vectors (the red dots on the left). The accelerations include not only a tangential acceleration but also an unnecessary centripetal acceleration.}
    \label{fig:2nd_derivative}
\end{figure}

The tangential component of joint motion on acceleration reading of the SU requires the measure of joint angular acceleration, which is not readily available in robot manipulators and therefore requires a numerical estimation.
%Although some robotic arms have torque sensors that allow the measurement of joint angular acceleration, many of them use an encoder to measure angular velocity instead. Therefore, the tangential acceleration needs to be numerically computed as shown in \cref{eq:pure_tangential_acceleration}.
Conventional methods (e.g. \cite{mittendorfer2012open}) employ the second derivative of the position vector ${}^{n}\vec{r}_{k}$ to estimate a SU's acceleration.
However, this differentiation does not include gravity and incorrectly incorporates an additional centripetal force from the SU moving on a rotating joint, as detailed in \cref{fig:2nd_derivative}.
This additional centripetal component is not only redundant, but also negatively affects optimization performance; therefore, in this work we remove non-tangential forces by employing \cref{eq: numerically_compute_acc}, followed by \cref{eq:pure_tangential_acceleration}:
% . On the right side, we show the resulting acceleration vectors in a 2D joint coordinate plane. It includes not only a tangential force but also an additional centripetal force which is not necessary. In our method, we removed those forces that are not tangential as in \cref{eq:pure_tangential_acceleration}.
%
\begin{align}
    \small
    \vec{a}_{\text{tan},k} &= \vec{e}_{\text{tan}} \cdot \frac{\vec{r}(h)+\vec{r}(-h)-2\cdot \vec{r}(0)}{h^2}
    \label{eq: numerically_compute_acc} \\
    ^{k}\vec{a}_{\text{tan},k} &= {}^{k}R_{n} \cdot \vec{a}_{\text{tan},k}
    \label{eq:pure_tangential_acceleration}
\end{align}
Above, $\vec{r}(h)$ is a parametrization of position vector ${}^{n}\vec{r}_{k}$ as function of discrete time $h=0.001$, and $\vec{e}_{\text{tan}}$ is an unit vector in the tangential direction.

\subsubsection{Error Functions}

For each SU$_k$ attached to joint $n$, we optimize its DH parameters $\Phi_k$ by separately minimizing two distinct error functions: the static acceleration error $E_{s}$ and the dynamic acceleration error $E_{d}$.
For each recorded static pose $p$, the $E_{s}$ only holds rotational information because it depends exclusively on gravitational acceleration; therefore, it can be used to estimate the rotational DH parameters included in ${}^{n}R_{k} (\Phi_k)$.
It is worth noting that there is a mis-alignment in coordinate frames: the measured acceleration ${}^{k}\vec{a}_{k,p}^{m}$ is defined in its own reference frame $k$, whereas the gravity vector ${}^{b}\vec{g}$ is defined in the base frame.
Therefore, the acceleration vector is converted into base frame using a rotation matrix ${}^{b}R_{n}$ as computed from forward kinematics. % ${}^{b}\vec{g}$ that includes ${}^{n}R_{k} (\Phi_k)$.
The static acceleration error is then defined as the L2 norm between the measured acceleration and the gravity vector:
\begin{equation}
    \small
    E_{s}(k) = \frac{1}{P} \sum_{p=1}^{P} \Bigl| {}^{b}R_{n} \cdot {}^{n}R_{k} (\Phi_k) \cdot {}^{k}\vec{a}_{k,p}^{m} - {}^{b}\vec{g} \Bigr|^2  ,
    \label{eq:static_error_function}
\end{equation}
where $P$ represents the total number of poses collected.
Conversely, the dynamic acceleration error $E_{d,k}$ is composed of both rotational and translational information.
Upon optimization of ${}^{n}R_{k} (\Phi_k)$ from \cref{eq:static_error_function}, the translational component of the problem can be then estimated via the L2 norm between the measured acceleration ${}^{k} \vec{a}_{k,d,p}^{m}$ and the ground truth estimated from kinematics ${}^{k} \vec{a}_{k,d,p} (\Phi_k)$ and expressed in the SU's reference frame:
\begin{align}
    \small
    \label{eq:dynamic_error_function}
    E_{d}(k) = \frac{1}{P} \sum_{p=1}^{P} \sum_{\substack{d>0, \\ d=n-2}}^{n}  \Bigl| {}^{k} \vec{a}_{k,d,p}^{m} - {}^{k} \vec{a}_{k,d,p} (\Phi_k) \Bigr|^2 ,
\end{align}
where $d$ represents the sole joint being actuated during the data collection process to exert an acceleration on each SU. Three previous joints before the $n$th joint are used to compute this error function. ${}^{k} \vec{a}_{k,d,p}$ can be computed from \cref{eq:total-acceleration} for joint $d$ and pose $p$. The optimization procedure is repeated for all SUs.

\section{Experiment Design}\label{sec:experiment}

\begin{figure}
    \centering
    \includegraphics[width=0.48\linewidth]{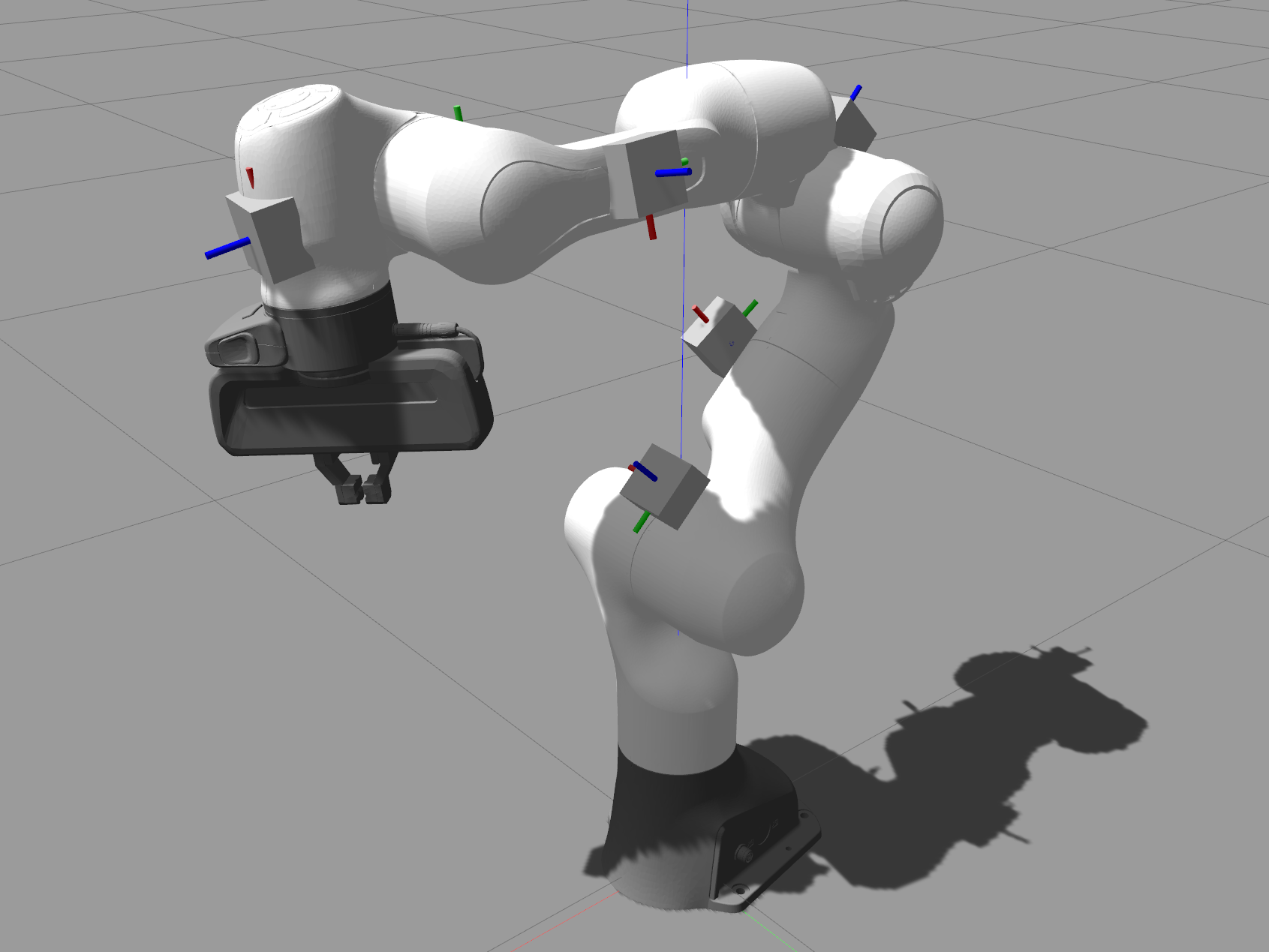}\
    \includegraphics[width=0.48\linewidth]{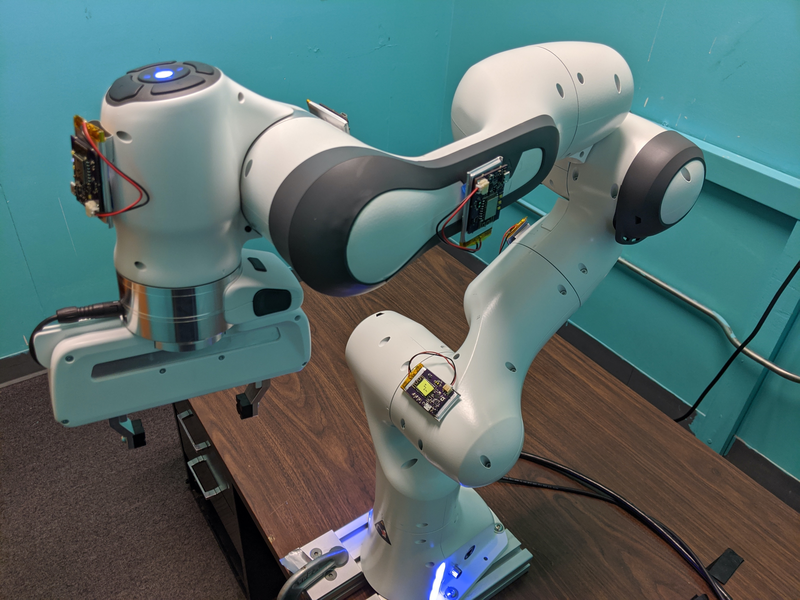}\\ \vskip 4pt
    \includegraphics[width=0.48\linewidth]{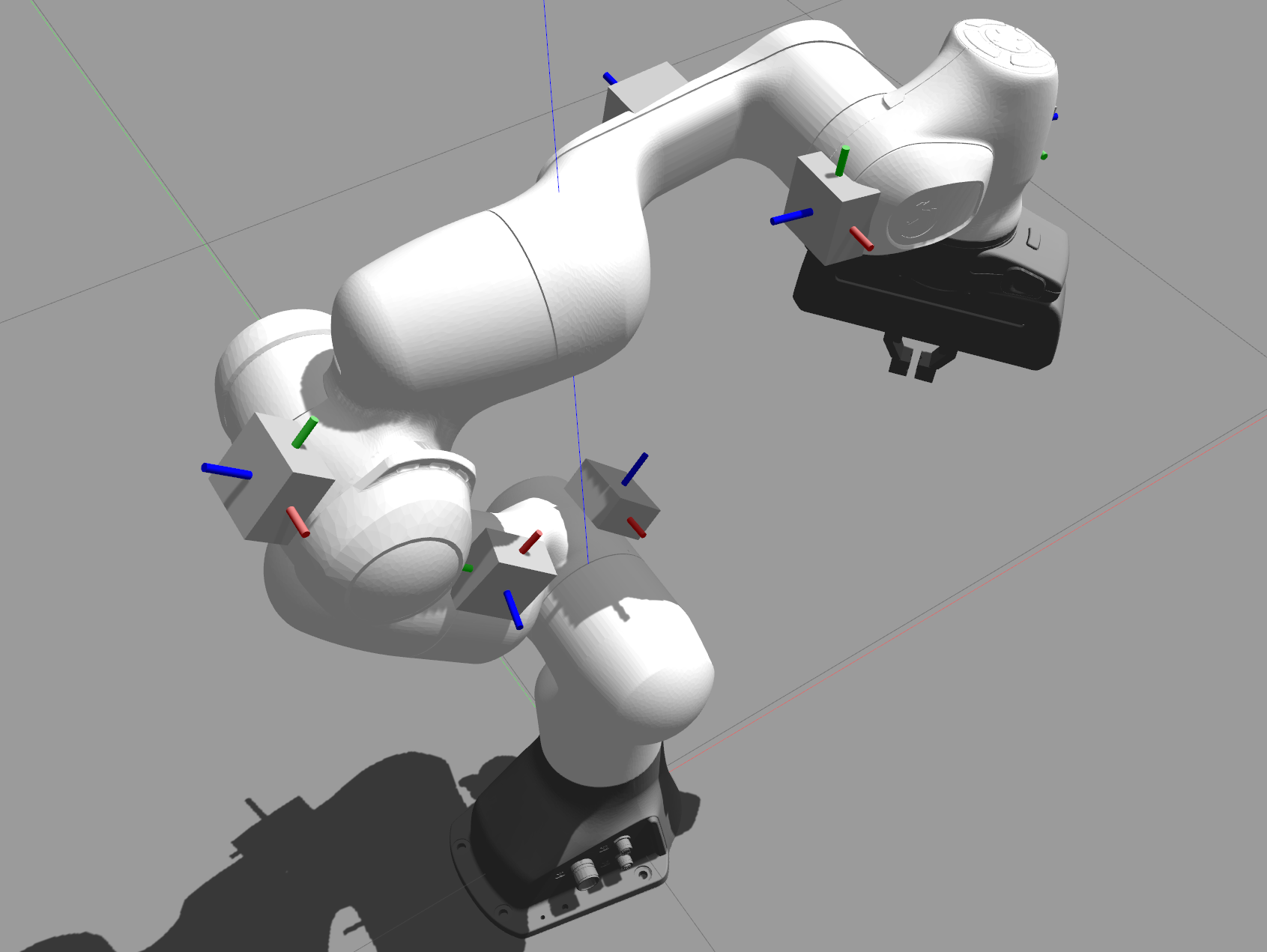}\
    \includegraphics[width=0.48\linewidth]{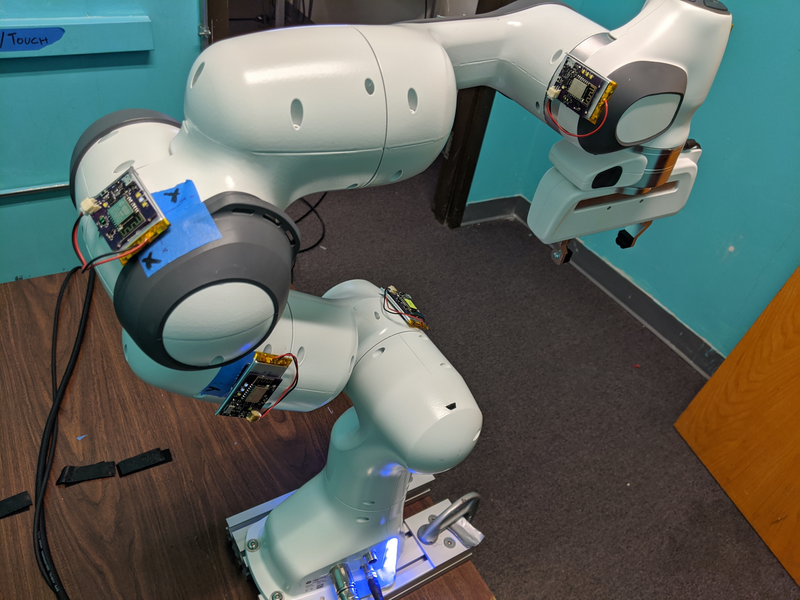}
        \caption{Comparison between estimated SU poses (left) and ground truth poses (right) on a Franka Emika Panda robot arm. The grey boxes represent the estimated SU with x (red), y (green), z (blue) axes respectively.}
    \label{fig:estimated_imus}
\end{figure}

To validate the proposed approach with reproducible and precise ground truth information, a quantitative evaluation is conducted in simulation via the Robot Operating System (ROS, \cite{quigley2009ros}) and the Gazebo simulator \cite{koenig2004design}.
A subsequent deployment on a real-world Franka Panda robot (see \cref{fig:panda,fig:estimated_imus}) is then used to provide qualitative analysis and a control example.
We demonstrate that our algorithm outperforms prior work \cite{mittendorfer2012open} and allows for precise autonomous calibration on 7-DoF robotic arms in both simulated and real-world environments.
In the following, we detail parameters of our experimental evaluation so as to improve reproducibility and repeatability of this work.
Data and algorithms are available at the associated GitHub repository\footnote{\ \url{https://github.com/HIRO-group/ros_robotic_skin}}.

Due to limitations of the original algorithm presented in \cite{mittendorfer2012open}, we modified it by including gravitational acceleration to follow \cref{eq:total-acceleration}.
For clarity, in the following we report a comparison of the proposed method against this enhanced version---which we hereinafter refer to as modified Mittendorfer's method (or \textsl{mMM}), as the original work was orders of magnitude worse than ours.
\begin{table*}
\centering
\caption{Comparison of the Average L2 Norm positional error [cm] and absolute distance in quaternion space (with units multiplied by $10^{-1}$ for reasons of space) of our method (OM) against the modified Mittendorfer method (mMM). $4$ different sets of poses are tested and optimization is run $10$ times per set; average and standard deviations of these $40$ trials are below.}
\label{tab:l2norm}
\begin{tabular}{r|r|c|c|c|c|c|c|c}
\multicolumn{2}{l|}{}              & SU$_1$ & SU$_2$ & SU$_3$ & SU$_4$ & SU$_5$ & SU$_6$ & Average                       \\ \hline
Positional & OM  & $0.28 \pm 0.15$                   & $0.39 \pm 0.15$                   & $0.78 \pm 0.39$                   & $1.15 \pm 0.88$                   & $0.80 \pm 0.36$                  & $0.25 \pm 0.095$                  & $\color{PineGreen}{0.66 \pm 0.60}$                        \\ %\cline{2-9}
Error [cm] & mMM & $5.3 \pm 7.1$                  & $2.6 \pm 0.23$                   & $5.4 \pm 4.9$                   & $9.5 \pm 9.0$                  & $1.4 \pm 0.91$                  & $2.3 \pm 2.4$                   & $\color{BrickRed}{4.4 \pm 5.9}$ \\ \hline
Quaternion & OM  & $0.10 \pm 0.058$                   & $0.054 \pm 0.041$                   & $0.023 \pm 0.012$                  & $0.019 \pm 0.013$                   & $0.021 \pm 0.023$                   & $0.045 \pm 0.053$                   & $\color{PineGreen}{0.044 \pm 0.059}$                        \\ %\cline{2-9}
Distance   & mMM & $5.5 \pm 5.9$                   & $0.16 \pm 0.11$                   & $3.6 \pm 6.1$                   & $0.057 \pm 0.022$                   & $0.044 \pm 0.015$                   & $2.5 \pm 4.3$                   & $\color{BrickRed}{2.0 \pm 4.4}$                        \\
\end{tabular}
\end{table*}
\begin{table*}
\centering
\caption{True and Estimated DH Parameters of one of the four SU configuration sets we tested.}
\label{tab:dhparam_comparison}
\begin{tabular}{l|l|l|l|l|l|l|l|l|l|l|l|l}
               & \multicolumn{2}{l|}{$j_1$ to SU$_1$} & \multicolumn{2}{l|}{$j_2$ to SU$_2$} & \multicolumn{2}{l|}{$j_3$ to SU$_3$} & \multicolumn{2}{l|}{$j_4$ to SU$_4$} & \multicolumn{2}{l|}{$j_5$ to SU$_5$} & \multicolumn{2}{l}{$j_6$ to SU$_6$} \\ \hline
               & true              & est              & true              & est              & true              & est              & true              & est              & true              & est              & true              & est              \\ \hline
$\theta_{v}$       & 1.571             & 1.571            & 0.000             & -0.014           & -1.571            & -1.571           & 3.141             & 3.141            & -1.571            & -1.571           & 1.571             & 1.565            \\ \hline
$d_{v}$              & 0.060             & 0.060            & -0.080            & -0.076           & 0.080             & 0.085            & -0.100            & -0.007           & 0.030             & 0.030            & 0.000             & 0.001            \\ \hline
$\theta_{SU}$              & -1.571            & -1.545           & 0.000             & -0.001           & 1.571             & 1.567            & 3.141             & 3.141            & 1.571             & 1.541            & -1.571            & -1.570           \\ \hline
$d_{SU}$          & 0.060             & 0.062            & 0.050             & 0.053            & 0.060             & 0.060            & 0.100             & 0.142            & 0.050             & 0.048            & 0.050             & 0.049            \\ \hline
$a_{SU}$ & 0.000             & 0.001            & 0.000             & 0.000            & 0.000             & 0.001            & 0.000             & -0.001           & 0.000             & -0.001           & 0.000             & -0.001           \\ \hline
$\alpha_{SU}$     & 1.571             & 1.583            & 1.571             & 1.568            & 1.571             & 1.570            & 1.571             & 1.565            & 1.571             & 1.570            & 1.571             & 1.571
\end{tabular}
\end{table*}
Similarly to \cite{mittendorfer2012open}, the base of the robot serves as the starting point for kinematic exploration; for simplicity, we aligned the base of the Panda arm to coincide with the world reference frame.
In both the simulation and the real world, we mount and simultaneously calibrate six SUs placed perpendicularly to the surface of each link---one per link except for the more proximal link.
It is worth noting that an SU placed on the first link cannot be calibrated because there exists an infinite number of solutions with the same acceleration values, due to the fact that all the dependent values $\vec{\dot{q}}$,  $\vec{\ddot{q}}$, and $\vec{r}$ are equal along the first link.
We used the native Gazebo IMU plugin to simulate the IMU physics and behavior.
The DH parameters are initialized randomly within reasonably large bounds: $\theta_v \in[-\pi;\pi],\ d_v \in[-1;1],\ \theta_{SU}\in[-\pi;\pi], \ d_{SU}\in[-1;1],\ a_{SU}\in[0;1], \ \alpha_{SU}\in[-\pi;\pi]$ (units are in meters for displacements and in radians for angles).

Data collection is performed only once both in simulation and in the real world. $P=16$ poses were collected, taking less than $20$ minutes. We used a joint velocity controller to oscillate at $2$rad/s. Both IMUs and controller messages were published and controlled at $100$Hz. Two seconds of rest were set between each sequence to prevent outliers and to make sure the robot could reach its desired position in time.

The parameter optimization step leveraged a randomized global optimization algorithm (MLSL\_LDS) \cite{kan1987newuoa} together with a derivative-free local optimizer (LN\_NEWUOA)
\cite{powell2004nloptnewuoa} of the NLopt library
 \cite{johnsonnlopt}.
Although we do initialize the DH parameters within a set of boundaries, we use NEWUOA unconstrained optimization to allow for better exploration of the DH parameter search space---which, in our case, is a $36$-dimensional space, due to each one of the six SUs having six parameters to optimize.
Due to the nature of the DH convention, there are multiple solutions for each SU's DH parameters, and using unconstrained methods allows us to more systematically explore those solutions and avoid local minima.
It is worth noting that performance of the proposed algorithm is increased when optimizing all IMUs concurrently, as data coming from multiple IMUs helps with redundancy and noise reduction; however, the proposed method works even when calibrating each IMU independently from the rest.
The optimization for the static acceleration error is terminated once the error is below a threshold of $0.01$ or if the parameter difference per iteration is less than $0.001$. Similarly, the optimization for the dynamic acceleration error is stopped once the change in the SU DH parameters reaches a threshold of less than $0.001$.
Both data collection and optimization were performed on a Lenovo Thinkpad X1 Extreme, 2nd Generation, with 12 Intel i7-9750H CPUs and a GeForce GTX 1650.

%!TEX root = ../root.tex
\section{Results and Discussion}\label{sec:results}

% \begin{figure*}
%     \centering
%     \includegraphics[width=\linewidth]{imgs/EstimatedPoses.pdf}
%         \caption{Comparison between estimated SU poses (first and third images from the left) and the ground truth SU poses (2nd and 4th images from the left) on Franka Emika Panda robot arm. The grey boxes represent the estimated SU with x (red), y (green), z (blue) axis respectively.}
%     \label{fig:estimated_imus}
% \end{figure*}

\subsection{Simulation experiments}
In simulation, we evaluate our algorithm based on the average L2 Euclidean distance and quaternion distance over all of the skin units.
To demonstrate the robustness of our approach, we randomized our experiment by the following:
 i) we tested $4$ different sets of SU configurations, i.e. we placed the SUs in $4$ different randomized locations along the robot's body;
ii) we optimized SU poses $10$ times per set.
Additional tests do not seem to alter average results nor to increase variance.
The mean and standard deviation of the resulting calibration on these $40$ trials are depicted in \cref{tab:l2norm}, and the corresponding estimated DH parameter values are shown in \cref{tab:dhparam_comparison}.
%Note that there are multiple solutions of DH parameters to represent one pose. We took those parameters that are aligned to one of the true DH parameters.
%
As mentioned in \cref{sec:experiment}, we chose not to compare against the original method in \cite{mittendorfer2012open} because of poor performance (about $90$cm positional error) and low convergence characteristics of the vanilla algorithm.
According to \cref{tab:l2norm}, while the modified version of Mittendorfer's method (mMM) had an average error of $4.4$cm and a high variance of $\pm 5.9$cm, our method (OM) has an error of $0.66$cm for the given SU placements, a six-fold improvement; similar considerations can be said of the orientation error.
In addition to the accuracy, we also validated the effectiveness of separating the original optimization problem into two parts.
Thanks to this solution, we measured a $4$x speed up of the separated problem ($433$ seconds) with respect to the original problem ($1797$ seconds).

\begin{figure*}
    \centering
    \includegraphics[width=\linewidth]{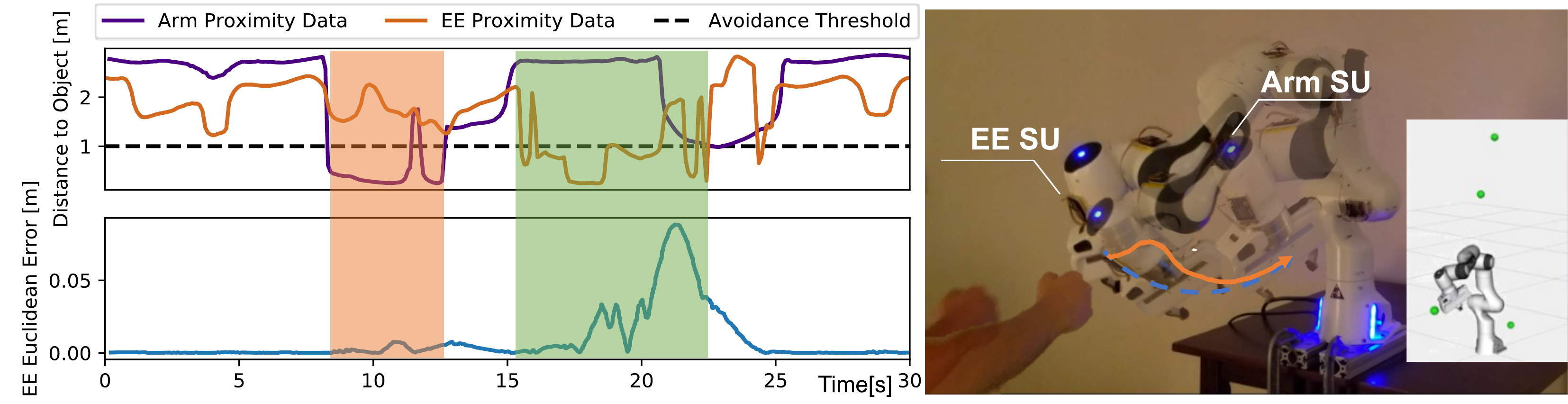}
    \caption{Trajectory redirection for obstacle avoidance.
    Right: a human approaches a robot with calibrated skin units; their accurate pose estimation allows the robot to precisely perceive the person from a distance.
    Bottom-Right: real-time visualization of the robot and obstacles (shown as green spheres) as sensed by the proximity sensors placed on the calibrated SUs; only below-threshold signals (arm and end-effector SU) affect the robot's motion.
    Left: combined graph of proximity data over time vs. the robot's end effector error along a circular trajectory. The two shaded areas indicate intervals when the human approaches a skin unit (below the avoidance threshold), causing only a temporary perturbation in the error. The perturbation stays low in the orange colored area (arm proximity activation) compared to the green area (end effector proximity activation) because the robot leverages the redundancy afforded by its kinematic chain to avoid colliding with the person while concurrently satisfying its main task objective.}
    \label{fig:controlexample}
\end{figure*}

\subsection{Real-world experiments}
To further test the effectiveness of the algorithm, we conducted an experiment in the real world---and more specifically on a Franka Panda robot.
The following is worth noting:
 i) the work in \cite{mittendorfer2012open} did not perform real-world calibration experiments, and to the best of our knowledge our paper is the first time such an approach has been validated on a real robot platform;
ii) we expect somewhat reduced performance due to the introduction of additional layers of complexity---from possible time delays in the robot network that misalign collected data, to sensor noise in the IMUs, to hardware limitations of the platform.
Similarly to the simulation, IMU data is retrieved at $100$Hz and recorded for $3$ seconds at each pose; however, we lengthen the resting time to $10$ seconds in order to allow for better separation of data collected at each pose.
Additionally, we modified the oscillation magnitude to $2.0$ and frequencies to $0.75$, $0.25$, $0.35$, $0.55$, $0.75$, $0.9$ and $0.9$Hz for each joint respectively, in order to exert high acceleration but restrain the displacement.
Data collection on the real robot was conducted on a computer with a real-time kernel and took less than $30$ minutes, to allow ample time for the real robot to move from one position to another.
We collected data for $P=12$ different poses.
A qualitative comparison of the actual SU poses against the estimated SU poses are is shown \cref{fig:estimated_imus}.
The grey boxes in simulation represent the estimated SUs and the corresponding pictures show the actual SU placements.
Although only qualitative, it is clear how the poses are, for the vast majority, very similar from one another---apart from one SU that seems to be more displaced apart.
On the real robot, the resulting average position error for all SUs was $5.1$cm and the average orientation error was $0.12$, whereas mMM achieves a $9.2$cm error for position and a $0.43$ error for orientation.
As mentioned in \cref{sec:experiment}, the original contribution from \cite{mittendorfer2012open} was significantly worse in the real world, with errors of up to $90$cm---i.e., an order of magnitude more than our work.
% The SUs colored in orange are estimated by our method and those colored in blue are estimated by enhanced Mittendorfer's method.
% Our estimated SU poses are close enough to the actual poses. However, the other estimated SU poses are far from the actual poses. The difference is significant that the SU poses estimated by their method cannot be utilized when in control.

\subsection{Control example}
After calibration, the estimated poses are then used for an obstacle avoidance control example, depicted in \cref{fig:controlexample} and shown in the accompanying video. 
To demonstrate the overall system, we implemented an obstacle avoidance controller to track waypoints in Cartesian space. We formalized it as a quadratic optimization problem to compute the optimal joint velocities while avoiding obstacles. To leverage the skin units for obstacle avoidance, we first find points on the robot's body closest to the obstacles and set constraints to restrict the approaching velocity. Further details can be found in \cite{escobedo2021contact}.
In the example in Figure \ref{fig:controlexample}, the robot is tasked with following a continuous circular trajectory in operational space, and constantly monitors the calibrated proximity readings at the nominal frequency of $50$Hz.
The bottom right image in \cref{fig:controlexample} shows a visualization of the robot and the obstacles detected in real time by the proximity sensors; the robot will take corrective actions if an obstacle is sensed within one meter of a skin unit (as depicted by the dashed line in the graph).
It is worth mentioning that in this demonstration we have chosen an arbitrary distance of one meter to guarantee operational safety in presence of humans; however, such threshold can be modified as function of task progress and perceived comfort of the human in presence of the robot.
The effectiveness of the control example is concretely demonstrated in the left graph.
The two shaded areas indicate when the human is approaching a skin unit below the avoidance threshold:
the perturbation stays low in the orange colored area (arm proximity activation), whereas it is more prominent in the green area (end effector proximity activation); in the first case, the robot effectively leverages its redundancy to avoid colliding with the person---which is not possible when the obstacle is approaching the end-effector of the robot's kinematic chain.
Our control example demonstrates that the estimation accuracy is sufficient for the motivations highlighted in \cref{sec:introduction}---that is, nearby space perception for close-proximity Human-Robot Collaboration.
The controller is able to quickly adjust its trajectory after a skin unit senses an obstacle, leading to operationally safe behavior. In addition to these results, we have shown further information on both the algorithm and the control example in the accompanying video (also available at this link \url{https://youtu.be/LzVkLmw5WA0}).

% This is the first ever presented video that exhibits the whole sequence of the kinematic calibration because the conventional method did not have any video nor open-sourced code.
% \ms{refer to the google sheets for all results - link is at https://docs.google.com/spreadsheets/d/1qFYWcEJta-c0bf6sjDq9hzT7d-IWrHRJRo3IOkN4Ug4/edit#gid=0}

% \input{5_discussion}
\section{Conclusions and Future Work}\label{sec:conclusions}

In this paper, we presented an accurate method for kinematically calibrating multiple skin units mounted on a collaborative robotic arm. Our approach is robust to noise and scales gracefully with the number of SUs, while not requiring expensive tracking systems with computational overhead.
The proposed work can effectively estimate SU poses with sub-cm precision in simulation, and less than $5$cm in the real world, for a total time from installation to effective utilization of less than $40$ minutes ($30$m for data collection $+\sim 7$m for optimization of multiple SUs).  
We have open-sourced our code and data for the research community, and we are hopeful that this demonstration will be the first of many to leverage calibrated skin unit poses for a future of plug-and-play distributed artificial tactile sensing for robotics.
%\href{https://github.com/HIRO-group/ros_robotic_skin}{https://github.com/HIRO-group/ros\_robotic\_skin}}.

In future work, we will focus our attention in four directions:
 i) extensively investigate the trade-off between  calibration accuracy and safe robot operation in human-populated environments (by comparing against human skin sensitivity and capability);
 ii) understand optimal placement of SUs along the robot's body to achieve best coverage and robust perception of the robot's surroundings;
 iii) further iterate on the artificial skin technology, adding heterogeneous sensing, flexible electronics, and dense whole-body coverage;
 iv) leverage i-iii) for novel robot behavior, such as environment-informed null-space control and human-aware motion planning. 
Ultimately, these applications will bring us closer to the realization of true, inherently safe human-robot interaction.

% \addtolength{\textheight}{12cm}   % This command serves to balance the column lengths

%%%%%%%%%%%%%%%%%%%%%%%%%%%%%%%%%%%%%%%%%%%%%%%%%%%%%%%%%%%%%%%%%%%%%%%%%%%%%%%%

%%%%%%%%%%%%%%%%%%%%%%%%%%%%%%%%%%%%%%%%%%%%%%%%%%%%%%%%%%%%%%%%%%%%%%%%%%%%%%%%

%%%%%%%%%%%%%%%%%%%%%%%%%%%%%%%%%%%%%%%%%%%%%%%%%%%%%%%%%%%%%%%%%%%%%%%%%%%%%%%%
% \input{notations}

% \input{appendix}

% \input{acknowledgment}

%%%%%%%%%%%%%%%%%%%%%%%%%%%%%%%%%%%%%%%%%%%%%%%%%%%%%%%%%%%%%%%%%%%%%%%%%%%%%%%%

\bibliographystyle{IEEEtran}
\bibliography{bibliography}

\end{document}